\documentclass[runningheads]{llncs}
\usepackage[T1]{fontenc}
\usepackage{graphicx}
\usepackage{url}
\usepackage{amsfonts}

\usepackage{amsmath}
\usepackage{amssymb}

\usepackage{algorithm}
\usepackage{algpseudocode}
\usepackage{hyperref}

\usepackage{svg}

\usepackage{enumitem}

\usepackage{xcolor}
\usepackage[table]{xcolor}
\usepackage{collcell}
\usepackage{array}
\usepackage{multirow}
\usepackage{todonotes}
\usepackage{amsmath}
\usepackage{amssymb}

\usepackage{caption}
\usepackage{subcaption}

\usepackage{booktabs}

\newcommand{\highlight}[1]{%
  \ifdim #1 pt > 0.05pt
    \textcolor{red}{#1}
  \else
    #1
  \fi
}

\newcolumntype{H}{>{\collectcell\highlight}c<{\endcollectcell}}

\begin{document}

\sloppy 

    %
    \title{Sampling on Random Subspaces under Limited Data in the
    Context of Exploratory Landscape Analysis}
    %
    \titlerunning{Sampling on Random Subspaces under Limited Data}
    %
    \author{Iv\'{a}n Olarte Rodr\'{\i}guez\inst{1}\orcidID{0009-0005-0748-9069}, 
    Anja Jankovic\inst{2}\orcidID{0000-0001-9267-4595}, 
    Thomas B\"{a}ck\inst{1}\orcidID{0000-0001-6768-1478}, \and
    Elena Raponi\inst{1}\orcidID{0000-0001-6841-7409}}

    \authorrunning{I.~Olarte Rodr\'{\i}guez et al.}

    \institute{
    Leiden Institute of Advanced Computer Science (LIACS), Leiden University, The Netherlands\\
    \email{\{i.olarte.rodriguez,t.h.w.baeck,e.raponi\}@liacs.leidenuniv.nl}
    \and
    RWTH Aachen University, Germany\\
    \email{jankovic@aim.rwth-aachen.de}
    }
    %

    \maketitle 
    \begin{abstract}
        Classical space-filling designs often fail to provide reliable statistical results for Exploratory Landscape Analysis (ELA) when only limited evaluation budgets are available, as commonly occurs in high-dimensional problems or other resource-constrained settings, 
        resulting in noisy and unstable landscape descriptors.

        To address this challenge, we propose an alternative sampling strategy for ELA based on random linear embeddings.
        Rather than sampling uniformly in the full decision space, we allocate the budget to randomly oriented low-dimensional subspaces and investigate whether this improves the robustness of the resulting landscape descriptors. 
        
        We compare full-space and embedding-based sampling strategies across several classical ELA feature sets on the noiseless Black-Box Optimization Benchmarking (BBOB) test suite from the COmparing Continuous Optimizers (COCO) environment, in a 20-dimensional setting.
        Our results suggest that random linear embeddings constitute a promising alternative for budget-constrained ELA, although their effectiveness remains dependent on the feature class and the underlying problem. 

    \end{abstract}
    \section{Introduction}
    \label{sec:Introduction}

    Deriving informative and robust descriptors of optimization problems is fundamental to many downstream tasks in numerical optimization, such as landscape similarity assessment, problem classification, and automated algorithm selection and configuration~\cite{jankovic2020landscape,kerschke_automated_2018,kerschke_automated_2019,kostovska2022per,MunEtAl2015}. In this context, Exploratory Landscape Analysis (ELA) \cite{mersmann_exploratory_2011} has established itself as a leading data-driven methodology. ELA transforms a collection of sampled decision vectors and their corresponding objective function values into a structured numerical feature space, thereby enabling the systematic characterization of black-box problems without relying on explicit analytical expressions.
 
    While ELA has demonstrated considerable promise, its applicability in expensive 
    settings remains limited \cite{olarte_rodriguez_does_2026,tanabe_towards_2021}. A central challenge is its reliance on relatively large sample sizes to obtain stable and reliable feature estimates~\cite{munoz_analyzing_2022,renau_expressiveness_2019}, which is an assumption that becomes impractical when function evaluations are costly \cite{tanabe_towards_2021}.
    This situation frequently arises in real-world applications, such as engineering design, where evaluating a single candidate solution may require computationally intensive simulations (\emph{e.g.,} finite element or computational fluid dynamics analyses) \cite{raponiKrigingassistedTopologyOptimization2019,ren_survey_2025}, or costly experimental procedures involving significant time and resources \cite{falaschetti_experimental_2025}. As a result, optimization is typically performed under strict evaluation budgets, where only a limited number of function evaluations are available to explore and characterize the search landscape.
    Under these conditions, commonly used ELA features exhibit substantial sampling variance, with values varying noticeably across independent samples of identical size~\cite{renau_exploratory_2020}. 

    As a result, the behavior of ELA features under limited budgets is inherently shaped by the sampling process. The sampling design not only determines the coverage of the decision space but also induces the neighborhood structure on which many feature classes rely. While space-filling designs promote global coverage, they lead to sparse representations in higher-dimensional settings, which reduces the stability and discriminative power of features based on local relationships \cite{kerschke_detecting_2015,renau_expressiveness_2019}. Conversely, sampling within defined regions of the landscape can improve the reliability of such features but introduces a bias toward specific regions of the search space. This highlights that the effectiveness of ELA features is governed by a trade-off between global representativeness and locally meaningful structure.
    
    To address this trade-off, we consider a sampling strategy based on random linear embeddings. Rather than projecting an already sampled set of points onto random subspaces, as in previous work~\cite{olarte_rodriguez_does_2026}, we consider the setting in which a fixed sampling budget is available and samples are generated directly within randomly oriented low-dimensional affine subspaces defined by linear embedding maps. This allows global variation to be explored along low-dimensional directions while maintaining higher sampling density within each subspace. Moreover, it avoids the artificial landscape distortions introduced by projection~\cite{olarte_rodriguez_does_2026}. As a result, neighborhood relationships are defined intrinsically within these subspaces, leading to more stable local feature estimates while still capturing variability in the ambient space, \emph{i.e.,} in the original representation.

    \paragraph{Our Contribution.}
    We study the stability of ELA features under limited sampling budgets by comparing features computed from samples in embedded subspaces and the full ambient space. The main contributions are:
    
    \begin{itemize}[leftmargin=*]
        \item We \textbf{propose a sampling strategy} based on randomly generated affine subspaces that reallocates a fixed evaluation budget to increase effective sampling density.
    
        \item We \textbf{conduct an empirical analysis across benchmark functions and feature sets}, distinguishing between feature computation in the ambient and reduced spaces, with the aim to examine how the effectiveness of subspace sampling depends on the objective function and feature class. 
    
        \item We \textbf{analyze the impact of compression ratios}, highlighting the trade-off between sampling density and information loss and its effect on feature stability.
    
        \item We \textbf{assess the preservation of ELA features} relative to a large-budget baseline and provide practical insights into the reliability of structured sampling under scarce evaluations.
    \end{itemize}
    
    Our results show that sampling in randomly embedded subspaces can improve the stability of subsets of ELA features under tight budgets. This provides a practical alternative to conventional space-filling designs, enabling more reliable landscape characterization in resource-constrained optimization problems.
    \paragraph{Code and Data Availability.}
    All code used to run the experiments and generate the results presented in this paper is publicly available at \url{https://github.com/olarterodriguezivan/Random_embeddings_BBOB}. A permanent archive of the datasets, results, and additional plots is available on Zenodo at \url{https://zenodo.org/records/19633768}.

    \section{Background}
    \label{sec:Background}

    \subsection{Sampling Challenges in Exploratory Landscape Analysis} 
    \label{subsec:ela_features}

    ELA~\cite{mersmann_exploratory_2011} provides
    a modular collection of feature sets designed to capture structural properties
    of optimization landscapes based on sampled data. These feature classes target
    complementary characteristics, such as global distributional properties, local
    neighborhood structure, meta-model behavior, and information-theoretic measures.
    
    Several ELA feature classes explicitly rely on local geometric or structural
    information. For example, \emph{information content} (\texttt{ic})
    features~\cite{munoz_exploratory_2015} estimate ruggedness and modality from
    symbol sequences derived from ordered function values along random walks.
    Similarly, \emph{nearest better clustering} (\texttt{nbc}) features~\cite{kerschke_detecting_2015,preuss_improved_2012}
    quantify funnel structures via directed graphs linking each point to its nearest improving neighbor.
    ELA relies on sufficiently dense, well-distributed samples to capture landscape characteristics, particularly local trends and neighborhood structure. Under sparse sampling (\emph{e.g.,} limited budgets or high dimensionality), feature estimates become unstable, exhibiting high variance across independent samples~\cite{renau_expressiveness_2019}. Consequently, the sampling strategy is critical: feature values depend strongly on how well samples represent the landscape and can vary significantly across sampling schemes~\cite{renau_exploratory_2020}.
    
    Recent work has explored trajectory-based (dynamic) sampling, where
    features are computed from points generated during an optimizer run~\cite{jankovic_towards_2021,kostovska2022per}.
    This avoids additional sampling cost and is effective for online algorithm selection.
    However, such samples are biased toward visited regions and are not space-filling,
    so the resulting features capture local, algorithm-dependent properties rather
    than global landscape characteristics. Consequently, they are not directly comparable
    to features obtained from uniform or quasi-random designs.

    \subsection{Random Linear Embeddings for Budget-Constrained Sampling}
    \label{subsec:RLE}

    Random linear embedding techniques \cite{art_owen_2013_mcbook,wang_bayesian_2016}~
    provide a structured way to restrict sampling to predefined low-dimensional subspaces of the ambient space. 

    The idea of randomized linear mappings comes from compressed sensing~\cite{candes_near-optimal_2006}, where signals can be recovered from a limited number of measurements when they admit sparse representations. In optimization, we use this idea in a broader sense: randomized linear mappings define transformed design variables that combine the original variables, allowing the objective function to be explored through a reduced parametrization without assuming sparsity in a prescribed basis.
    
    This principle already underlies several high-dimensional Bayesian optimization methods, including REMBO~\cite{wang_bayesian_2016}, ALEBO~\cite{Letham2020Re}, and PCA-based approaches~\cite{antonov_high_2022,raponi_high_2020}. The same reasoning applies to ELA, where the goal is to characterize landscape structure through compact numerical descriptors. Under a limited sampling budget, samples generated from such transformed variables can provide a denser and more reliable characterization of relevant landscape features than samples distributed uniformly across the full ambient domain.

    Building on this principle, the present work systematically investigates the stability of ELA features when sampling is performed on random linear embeddings. We compare feature estimates obtained \textit{(i)}~from embedded subspaces and \textit{(ii)}~from uniform sampling in the ambient space. This comparison allows us to assess whether embedding-based sampling can improve the robustness of landscape descriptors under strict evaluation budgets.
    In contrast to uniformly distributing points across the entire decision domain, embedding-based sampling emphasizes structured coverage of linear subspaces. From the perspective of landscape analysis, this may provide more reliable neighborhood relations and graph structures when evaluation budgets are limited, thereby potentially stabilizing feature classes that rely on local structural information.


    \section{Methodology}
    \label{sec:methodology}
    

    To address the limitations of space-filling sampling under restricted evaluation budgets, in this work we adopt a sampling strategy for ELA based on random linear embeddings. Rather than distributing samples across the full search space, we allocate them within low-dimensional subspaces.
    This increases the effective sampling density and improves local resolution, which is critical for estimating neighborhood relationships and local structural properties. As a result, embedding-based sampling can provide more reliable geometric and graph-based structures under limited budgets, thereby improving the stability of ELA features that depend on local information.
    

    \begin{figure}[tb]
        \centering
        \includegraphics[width=\linewidth]{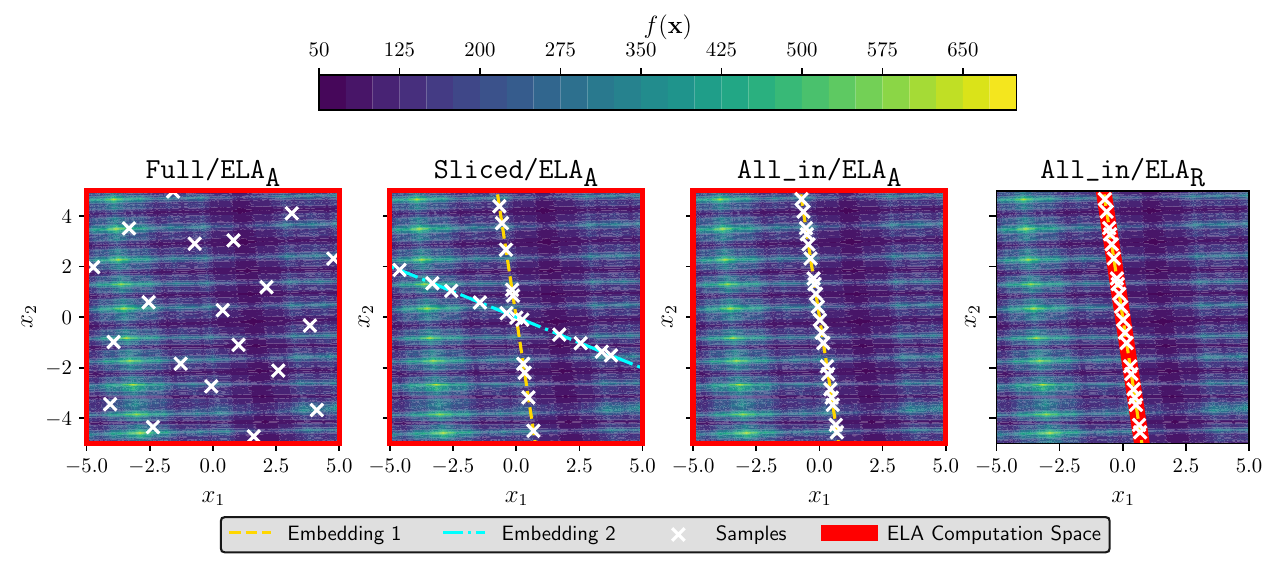}
        \caption{Representation of the proposed sampling strategies for ELA. All the panels show the original 2D landscape of the Weierstrass (f16) function with points sampled on the landscape. Each panel depicts different sampling strategies. The \textcolor{red}{red} lines represent the space wherein ELA features are computed.
        }
        \label{fig:ela_sampling_rep}
    \end{figure}

    To generate samples and compute the ELA features afterwards, we consider four sampling strategies (graphical representation in Fig. \ref{fig:ela_sampling_rep}):
    %
    \begin{enumerate}[leftmargin=*, label=(\arabic*)]
        \item \textbf{Full-space sampling:} 
        Samples are drawn uniformly in the full $D$-dimensional decision space $\mathcal{X}= [l,u]^{D}$, and ELA features are computed directly in this space. 
        We denote this setting by \texttt{Full/ELA$_{\texttt{A}}$}, where \texttt{A} indicates evaluation in the ambient space.
        \item \textbf{Multiple random embeddings (distributed budget):} 
        We generate $K$ random linear embeddings of effective dimension $d$ with compression ratio $r = d/D$. 
        The total sampling budget $N_s$ is evenly distributed, yielding $N_s/K$ samples per embedding, where $K = \lfloor 1/r \rfloor$. 
        All sampled points are mapped back to the $D$-dimensional space, and ELA features are computed on their union. 
        This approach is inspired by random projection techniques from compressed sensing \cite{candes_dantzig_2007}. 
        We refer to it as \texttt{Sliced/ELA$_{\texttt{A}}$}, reflecting the use of multiple embeddings (``slices'') with evaluation in the ambient space.
        \item \textbf{Single embedding (ambient evaluation):} 
        The full sampling budget $N_s$ is allocated to a single $d$-dimensional random embedding. Samples are mapped to the $D$-dimensional space, where ELA features are computed. 
        By analogy to poker, we denote this strategy as \texttt{All\_in/ELA$_{\texttt{A}}$}, indicating that the entire budget is committed to one embedding with evaluation in the ambient space.
        \item \textbf{Single embedding (reduced evaluation):} 
        As above, all samples are generated within a single $d$-dimensional embedding, but ELA features are computed directly in the reduced space. 
        This variant is denoted \texttt{All\_in/ELA$_{\texttt{R}}$}, where \texttt{R} indicates evaluation in the reduced space.
    \end{enumerate}

    The distinction between strategies (3) and (4) is important because linear embeddings can introduce strong linear dependencies between variables when represented in the ambient space. 
    This can negatively affect feature sets
    that rely on regression or classification models (\emph{e.g.,} \emph{meta-model} or \emph{level-set} features), potentially leading to ill-conditioned estimation problems. 
    Algorithm~\ref{alg:embedded_sampling} summarizes the general procedure for the computation of ELA features based on embedding-based sampling.
    The method consists of three main components:
    \textit{(i)} an \textbf{embedding generation step} (Algorithm~\ref{alg:generate_embeddings});
    \textit{(ii)} a \textbf{sampling step} performed in the embedded space,
    followed by \textbf{evaluation of the objective function} in the ambient space (Algorithm~\ref{alg:sample_and_evaluate}); and
    \textit{(iii)} a \textbf{feature computation step}, controlled by a flag indicating whether features should be computed in the embedded space (Algorithm~\ref{alg:compute_features}). Symbols are introduced progressively within the corresponding pseudocode blocks, where they become relevant.
    
    %
    %
    \begin{footnotesize}
        \begin{algorithm}[b]
            \caption{Embedded Sampling and Feature Computation Pipeline}
            \label{alg:embedded_sampling}
            \begin{algorithmic}[1]
                        
            \Require 
            Problem dimension $D$, embedding dimension $d$, number of subspaces $K$, sampling budget $N_s$, sampling strategy $\mathcal{S}$, search domain $\mathcal{X}$, objective function $f$, and flags: $\texttt{normalize}$, $\texttt{compute\_in\_reduced\_space}$ 
            
            \Ensure Sampled points $\{\mathbf{x}_i, \mathbf{z}_i, y_i\}_{i=1}^{N_s}$ and ELA feature vectors $\mathbf{t}\text{ (computed in ambient space)}$, $\mathbf{t}' \text{ (computed in embedded space)}$ 
            
            \State $\{\mathbf{A}^{(k)}, {\mathbf{A}^{\dagger}}^{(k)}\}_{k=1}^K \gets$
            \Call{GenerateEmbeddings}{$d, D, K, \texttt{normalize}$} 
            
            \State $\{(\mathbf{x}_i, \mathbf{z}_i, y_i)\}_{i=1}^{N_s} \gets$
            \Call{SampleAndEvaluate}{$\mathbf{A}^{(k)}, {\mathbf{A}^{\dagger}}^{(k)}, N_s, \mathcal{S}, \mathcal{X}, f$}
            
            \State $(\mathbf{t}, \mathbf{t}') \gets$
            \Call{ComputeFeatures}{$\{\mathbf{x}_i, \mathbf{z}_i, y_i\}, \texttt{compute\_in\_reduced\_space}$}
            \end{algorithmic}
        \end{algorithm}
    \end{footnotesize}
    %
    %
    In Algorithm~\ref{alg:generate_embeddings}, the main component is the construction of random linear embeddings using the matrix $\mathbf{A} \in \mathbb{R}^{D \times d} $. 
    The entries of $\mathbf{A}$ are sampled independently from a standard normal distribution, \emph{i.e.,} $\mathcal{N}(0,1)$. Optionally, a normalization step, controlled by the flag \texttt{normalize}, is applied, in which each column of $\mathbf{A}$ is scaled to unit norm. This step helps preserve Euclidean distances between the embedded and ambient spaces, which is important since ELA features are not invariant to such distortions.
  
    Next, the pseudo-inverse of \(\mathbf{A}\), denoted by \(\mathbf{A}^{\dagger}\), is computed. This step is required because points generated in the embedded space may be mapped, through the inverse transformation, to points that lie outside the feasible domain of the original ambient problem. To address this issue, Algorithm~\ref{alg:sample_and_evaluate} first draws samples in the embedded space according to a sampling strategy \(\mathcal{S}\), such as Sobol, Latin hypercube, or Halton sampling. Since these sampling strategies generate points in the unit hypercube \([0,1]^d\), the samples must be rescaled to the search region defined in the embedded space. In this work, the embedded search box is scaled as
    \(
    \frac{1}{\sqrt{r}} [l,u]^d
    \). This scaling accounts for the relative change in the length of the maximum diagonal of the unit hypercube when moving from the ambient dimension \(D\) to the embedding dimension \(d\). After the embedded samples are mapped back to the ambient space, the resulting points are clipped onto the feasible domain using the operator \(\Pi_{\mathcal{X}}(\cdot)\), as shown in line~6. This step ensures that the final candidate points satisfy the bounds of the ambient space \(\mathcal{X}=[l,u]^{D}\).

    However, since some feature computations require points to be expressed in the embedded space, an additional step is necessary. After correction, the feasible points in the ambient space are mapped back to the embedded space using the pseudo-inverse $\mathbf{A}^{\dagger}$. This ensures consistency between the domain in which feasibility is enforced and the domain in which features are evaluated. Note that clipping in the previous step may move points away from the embedded subspace and concentrate them near the domain boundary. Since our aim is to characterize the landscape from feasible samples, rather than to model and optimize the objective directly in the embedded space, we treat this distortion as negligible for feature computation.
    \begin{footnotesize}    
        \begin{algorithm}[h!]
            \caption{GenerateEmbeddings}
            \label{alg:generate_embeddings}
            \begin{algorithmic}[1]
            \Require Problem dimension $D$, embedding dimension $d$, number of subspaces $K$, and a flag: $\texttt{normalize}$
            \Ensure Embedding matrices $\{\mathbf{A}^{(k)}, {\mathbf{A}^{\dagger}}^{(k)}\}_{k=1}^K$
            
            \For{$k=1,\dots,K$}
                \State Sample $\mathbf{A}^{(k)} \in \mathbb{R}^{D \times d}$ \Comment{Elements are sampled according to \(\mathcal{N}(0,1)\)}
            
                \If{$\texttt{normalize}$} 
                    \For{$j=1,\dots,d$}
                        \State $\mathbf{A}^{(k)}_{:,j} \gets \mathbf{A}^{(k)}_{:,j} / \|\mathbf{A}^{(k)}_{:,j}\|_2$ \Comment{Column-wise normalization}
                    \EndFor
                \EndIf
            
                \State ${\mathbf{A}^{\dagger}}^{(k)} \gets \texttt{pinv}(\mathbf{A}^{(k)})$ \Comment{Compute Moore-Penrose pseudo-inverse \cite{penrose_best_1956}}
            \EndFor
            \end{algorithmic}
        \end{algorithm}
    \end{footnotesize}
    \begin{footnotesize}
        \begin{algorithm}[h!]
            \caption{SampleAndEvaluate}
            \label{alg:sample_and_evaluate}
            \begin{algorithmic}[1]
            \Require Random projection matrices and their pseudo-inverses $\{\mathbf{A}^{(k)}, {\mathbf{A}^{\dagger}}^{(k)}\}$, sampling budget $N_s$, sampling strategy $\mathcal{S}$, search domain $\mathcal{X}$, objective function $f$
            \Ensure Sampled points $\{(\mathbf{x}_i, \mathbf{z}_i, y_i)\}_{i=1}^{N_s}$
            
            \State $N_k \gets \lfloor N_s / K \rfloor$ \Comment{Determine number of samples per embedding}
            
            \For{$k=1,\dots,K$}
                \For{$i=1,\dots,N_k$}
                    \State $\mathbf{z}_i \gets \mathcal{S}(d)$ \Comment{Sample in \(1/{\sqrt{r}} \cdot [l,u]^{d}\)}
                    \State $\mathbf{x}_i \gets \mathbf{A}^{(k)} \mathbf{z}_i$ \Comment{Express sample in ambient space coordinates}
                    \State $\mathbf{x}_i \gets \Pi_{\mathcal{X}}(\mathbf{x}_i)$ \Comment{Clip onto box $\mathcal{X}= [l,u]^{D}$}
                    \State $y_i \gets f(\mathbf{x}_i)$ \Comment{Evaluate the objective  function}
                    \State $\mathbf{z}_i \gets {\mathbf{A}^{\dagger}}^{(k)} \mathbf{x}_i$ 
                    \Comment{Map clipped point back to reduced-space coordinates}                   
                \EndFor
            \EndFor
            \end{algorithmic}
        \end{algorithm}
    \end{footnotesize}
    \begin{footnotesize}
    	\begin{algorithm}[h!]
    		\caption{ComputeFeatures}
    		\label{alg:compute_features}
    		\begin{algorithmic}[1]
    			\Require Sampled points $\{(\mathbf{x}_i, \mathbf{z}_i, y_i)\}_{i=1}^{N_s}$, with a flag: $\texttt{compute\_in\_reduced\_space}$
    			\Ensure ELA feature vectors $\mathbf{t} \text{ (computed in ambient space)}$, $\mathbf{t}' \text{ (computed in embedded space)}$
    			
    			\If{$\texttt{compute\_in\_reduced\_space}$}
    			\State $\mathbf{t}' \gets \mathcal{F}\left(\{ (\mathbf{z}_i, y_i)\}_{i=1}^{N_s} \right)$
    			\EndIf
    			
    			\State $\mathbf{t} \gets \mathcal{F} \left (\{(\mathbf{x}_i, y_i)\}_{i=1}^{N_s} \right)$
    		\end{algorithmic}
    	\end{algorithm}
    \end{footnotesize}

    Finally, ELA features are computed once the sample points are represented in the ambient space and their corresponding function values have been evaluated. Algorithm~\ref{alg:compute_features} outlines the standard procedure for computing the feature vector $\mathbf{t}$ via the operator $\mathcal{F}[\cdot]$. An additional flag (\texttt{compute\_in\_reduced\_space}) enables the computation of an alternative feature vector $\mathbf{t}'$ in the reduced space when required.
    %

    \section{Experimental Setup}
    \label{sec:experimental_setup}
    \paragraph{Benchmark Problems.}
    We evaluate the proposed sampling strategies on noiseless functions from the Black-Box Optimization Benchmarking (BBOB) test suite from the COmparing Continuous Optimizers (COCO) environment \cite{hansen_coco_2021}. This suite comprises 24 continuous optimization problems that cover a wide range of landscape characteristics, including separability, conditioning, and multimodality. 
    The search domain of the BBOB functions is defined over $[-5,5]^D$. 
    Function evaluations and instance definitions are obtained using \texttt{IOHexperimenter} \cite{IOHexperimenter}, for which the first 15 instances of each function are used.
    Experiments are conducted only for $D = 20$, representing a moderate-dimensional setting \cite{tanabe_towards_2021}.
    
    \paragraph{Sampling Budget and Strategies.}
    %
    For each problem, the total sampling budget is fixed to
    \(
    N_{s} = 10 \cdot D
    \).
    Samples are generated using the four strategies described in Section~\ref{sec:methodology}.
    
    For embedding-based methods, the reduced dimension is determined by the compression ratio
    \(
    r = d/D,\; r \in \{0.1, 0.25, 0.5\}
    \).
    In the multi-embedding setting, the number of embeddings is set to
    \(
    K = \lfloor 1/r \rfloor
    \),
    with the budget uniformly allocated across embeddings. This yields \(K \in \{10,4,2\}\) embeddings with \(\{20,40,100\}\) samples per embedding for the respective values of \(r\).
    
    Sampling is performed via Latin Hypercube Sampling (LHS) in both ambient and embedded spaces. The search domain in the embedded space is expanded to \(1/{\sqrt{r}} \cdot [-5,5]^d\). 
    This was performed with the \texttt{scipy.stats.qmc} package \cite{2020SciPy-NMeth}. 
    
    Each configuration is repeated 40 times with different random seeds. For the \texttt{slice} approach, each embedding is generated using distinct seeds to avoid repetition. 
        %


    \paragraph{Considered Features}
    We consider six classical ELA feature sets: $y$-\emph{distribution} (\texttt{ela\_distr}), \emph{level-set} (\texttt{ela\_level}), \emph{meta-model} (\texttt{ela\_meta}), \emph{nearest-better clustering} (\texttt{nbc}), \emph{dispersion} (\texttt{disp}), and \emph{information content} (\texttt{ic}). In addition, \emph{fitness-distance correlation} (\texttt{fitness\_distance}) is included solely as a baseline, as its computation requires knowledge of the global optimum, which is typically unavailable in practice.
    
    We exclude the \texttt{pca} feature set \cite{kerschke_comprehensive_2017} as it primarily captures the geometric distribution of sampled points in the search space and is thus strongly influenced by the sampling strategy rather than intrinsic landscape properties \cite{long_bbob_2022}.
    
    The selected feature sets are computationally inexpensive \cite{jankovic2020landscape,kostovska2022per} and can be computed from a single sample without requiring repeated evaluations or structured sampling. An overview is given in Table~\ref{tab:feature_classes}, with further details available in the cited literature.
    \begin{table}[th]
        \centering
        \caption{Feature sets considered in this study.}
        \label{tab:feature_classes}
        \renewcommand{\arraystretch}{0.90}
        \begin{footnotesize}
            \begin{tabular}{llc}
                \toprule \textbf{Feature set} & \textbf{Name}                                                                 & \textbf{Num. features} \\
                \midrule \texttt{ela\_distr}  & distribution-based features \cite{mersmann_exploratory_2011}                  & 3                      \\
                \texttt{ela\_level}           & level-set features \cite{mersmann_exploratory_2011}                           & 9                      \\
                \texttt{ela\_meta}            & meta-model features \cite{mersmann_exploratory_2011}                          & 9                      \\
                \texttt{nbc}                  & nearest-better clustering \cite{kerschke_detecting_2015,preuss_improved_2012} & 5                      \\
                \texttt{disp}                 & dispersion features \cite{lunacek_dispersion_2006}                            & 16                     \\
                \texttt{ic}                   & information content features \cite{munoz_exploratory_2015}                    & 5                      \\
                \texttt{fitness-distance}     & fitness distance correlation \cite{jones_fdc_1995}                            & 6                      \\
                \bottomrule
            \end{tabular}%
         \end{footnotesize}
    \end{table}
    \paragraph{Feature Computation.}
    ELA features are computed using the \texttt{pflacco} package \cite{prager_pflacco_2024}. 
    Depending on the strategy, features are computed either in the ambient space or directly in the reduced subspace as explained in Section \ref{sec:methodology}.
    \paragraph{Evaluation Protocol.}
    We evaluate sampling strategies by measuring how well feature distributions obtained with a reduced budget approximate a high-fidelity reference.
    For each ELA feature, function, and instance, a reference distribution is computed using $2000$ ($100 \cdot D$) LHS samples. In line with Renau et al.~\cite{renau_expressiveness_2019}, these values are not interpreted as exact feature quantities, but rather as high-budget approximations within a fixed sampling scheme. The resulting distribution serves as an operational baseline.
    
    \paragraph{Distance measure.}

    The discrepancy between reference and approximated distributions is quantified using the Wasserstein-1 (Earth Mover’s) distance \cite{ramdas_wasserstein_2015}, computed per feature–function–instance triplet. This metric compares empirical distributions directly and is sensitive to differences across their full support. In contrast to divergence-based measures such as the Kullback-Leibler divergence, it remains well-defined under limited or non-overlapping support, providing a stable notion of distributional discrepancy without relying on parametric or moment-based assumptions.
    
    
    \paragraph{Aggregation and comparison.}
    For each feature and function, the Wasserstein distances are aggregated over instances and used to rank the sampling strategies based on the median.
    Rankings are averaged across \textit{(i)} ELA features and \textit{(ii)} functions to analyze overall performance. Results are reported using average ranking plots and Critical Difference (CD) diagrams \cite{demsar_statistical_2006}, where statistical significance is assessed via the post-hoc Nemenyi test \cite{nemenyi1979distribution}. 


    \section{Results}
    \label{sec:Results}
\begin{figure}[tb]
    \centering
    \includegraphics[width=\linewidth]{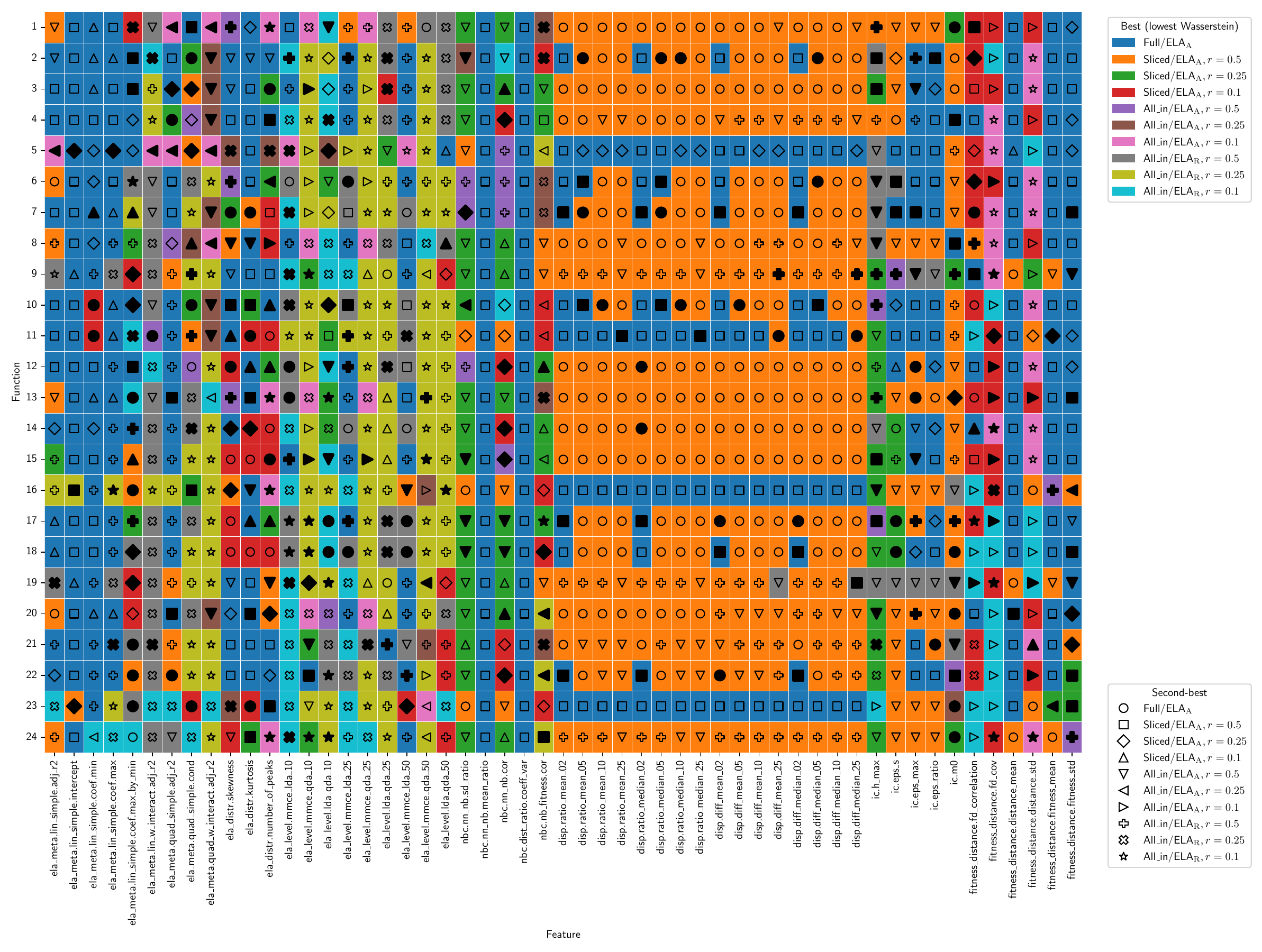}
    \caption{Rankings of ELA feature sampling methods based on the Wasserstein-1 distance relative to LHS sampling with 2000 $(100 \cdot D)$ samples. Colored blocks indicate the method that achieved the lowest median distance for each function--feature pair, while the symbol overlaid on each block denotes the second-lowest distance. A \textbf{filled} symbol indicates that the difference between the methods with the lowest and second-lowest distances is not statistically significant with $95\%$ confidence according to the signed Wilcoxon test.}
    \label{fig:wasserstein_rankings}
\end{figure}

    Following the experimental setup choices described in Section~\ref{sec:experimental_setup}, we summarize the ranking and distance measures in a compact form in Figure~\ref{fig:wasserstein_rankings}.
    In particular, we empirically evaluate ten different methods, constructed based on the four strategies outlined in Section~\ref{sec:methodology}; we use \texttt{Full/ELA$_{\texttt{A}}$} as a baseline, and we assess the performance of the combinations of three embedding-based strategies (\texttt{Sliced/ELA$_{\texttt{A}}$}, \texttt{All\_in/ELA$_{\texttt{A}}$}, 
    \texttt{All\_in/ELA$_{\texttt{R}}$}) with three different values of the compression ratio $r$. 
    Figure~\ref{fig:wasserstein_rankings} presents a heatmap of method rankings based on the Wasserstein distance between estimated and reference distributions of ELA features across the BBOB test suite. Rows correspond to functions and columns to features; colors indicate the best-performing method, while symbols denote the second-best.


    Overall, the heatmap does not reveal strong global trends. Nevertheless, two consistent observations can be made. First, \texttt{Full/ELA$_{\texttt{A}}$} tends to outperform subspace-based approaches for the subclass \texttt{ela\_meta.lin\_simple}. This is expected, as meta-model-based features aim to capture global landscape structure, which may not be preserved under projection to lower-dimensional subspaces.
    Second, \texttt{Sliced/ELA$_{\texttt{A}}$} with $r=0.5$ frequently achieves the best ranking for the \texttt{disp} features. This indicates that these features can often be estimated reliably from subspace samples, suggesting a certain robustness to localized sampling. 
    However, this behavior is not consistent across all functions. In particular, for functions exhibiting strong anisotropy or ill-conditioning, such as the Discus function (f11), as well as highly irregular landscapes such as the Weierstrass (f16) and Katsuura (f23) functions, subspace sampling appears insufficient to capture relevant structural properties. In these cases, full-space sampling provides more accurate approximations of the feature distributions. 

    For the \texttt{ela\_level} features, the best-performing strategies are consistently of the \texttt{All\_in/ELA$_{\texttt{R}}$} type, with $r=0.25$ often achieving the top ranking. While no single configuration dominates uniformly, concentrating samples within a single subspace is preferred over full-space sampling. In fact, \texttt{Full/ELA$_{\texttt{A}}$} is rarely among the top strategies, even as a second-best choice, reinforcing that it is not well suited for these features. The variability observed across functions reflects sensitivity to the compression ratio rather than to the sampling paradigm itself. The remaining question is how far the compression ratio can be reduced without losing essential structure.

\begin{figure}[bt]
        \centering
        \begin{subfigure}[t]{0.42\linewidth}
            \includegraphics[trim=0.15cm 0cm 4.9cm 0.58cm, clip,width=\linewidth]{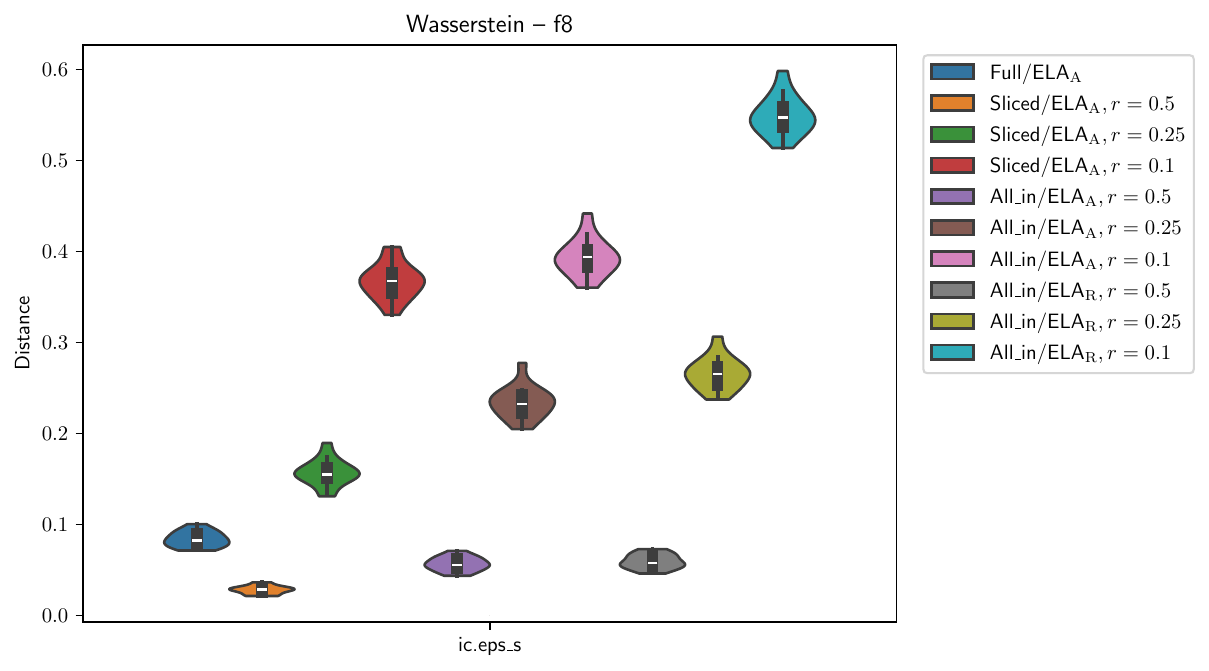}
            \caption{Rosenbrock (f8)}
        \end{subfigure}
        \hfill
        \begin{subfigure}[t]{0.42\linewidth}
            \includegraphics[trim=0.15cm 0cm 4.9cm 0.58cm, clip,width=\linewidth]{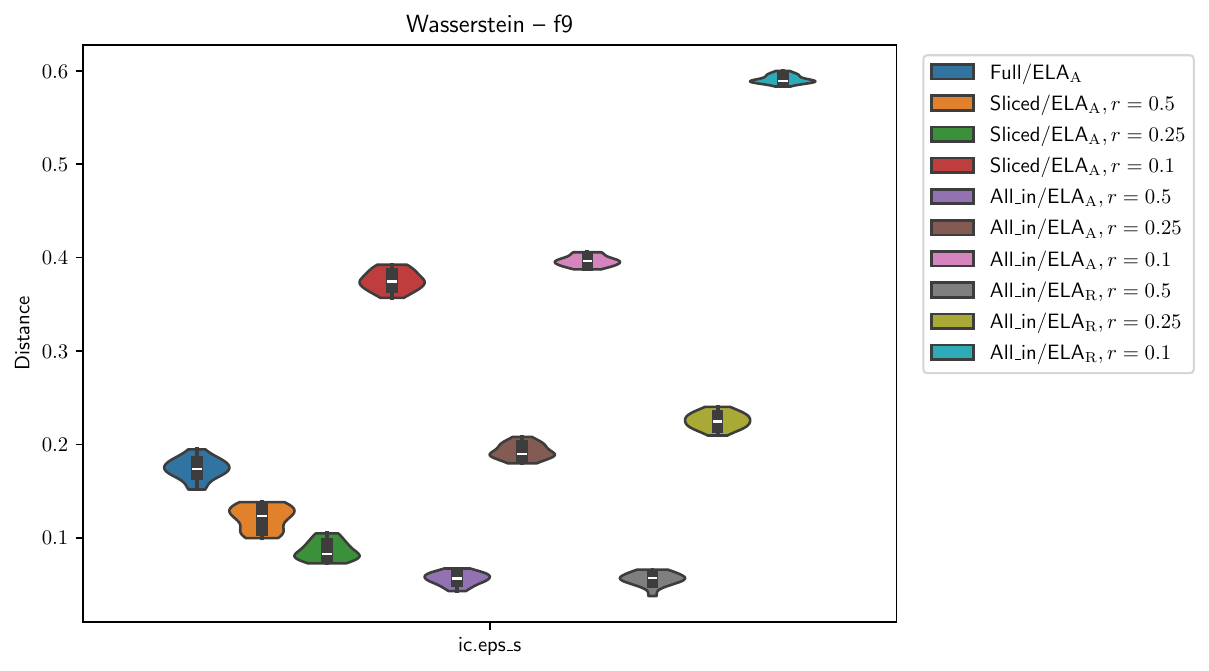}
            \caption{Rotated Rosenbrock (f9)}
        \end{subfigure}
        \hfill
        \begin{subfigure}[t]{0.13\linewidth}
            \includegraphics[trim=15.5cm 5cm 0cm 0.9cm, clip,width=\linewidth]{Images/eps_s_f9_violin_wasserstein.pdf}
        \end{subfigure}
        
        \caption{Per instance Wasserstein distance distribution of the \texttt{ic.eps\_s}.}
        \label{fig:eps_s_per_instance_wasserstein}
    \end{figure}

    To illustrate how Wasserstein distances vary depending on both the function and the analyzed feature, we present Figure~\ref{fig:eps_s_per_instance_wasserstein}, which shows distributions of the Wasserstein distances across instances for the Rosenbrock (f8) and Rotated Rosenbrock (f9) functions for the \texttt{ic.eps\_s} feature, which defines the sensitivity of the landscape to variable scaling and, intuitively, quantifies how much of the landscape structure is preserved when the resolution is changed from fine- to coarse-grained. 
    This example highlights that, despite the strong similarity between these functions, noticeable differences in Wasserstein distances can still arise. This observation motivates the use of aggregate statistics such as medians and rankings. 

    Because Wasserstein distances are on different scales for different features and functions, \emph{i.e.,} reflecting differences in feature magnitude and variability rather than in method quality, direct comparisons of raw Wasserstein distance values across features or functions would be misleading. 
    We therefore conduct a ranking-based analysis to identify relative strengths and weaknesses of the sampling strategies.
    \begin{figure}[t]
        \centering
        \includegraphics[width=0.975\linewidth]{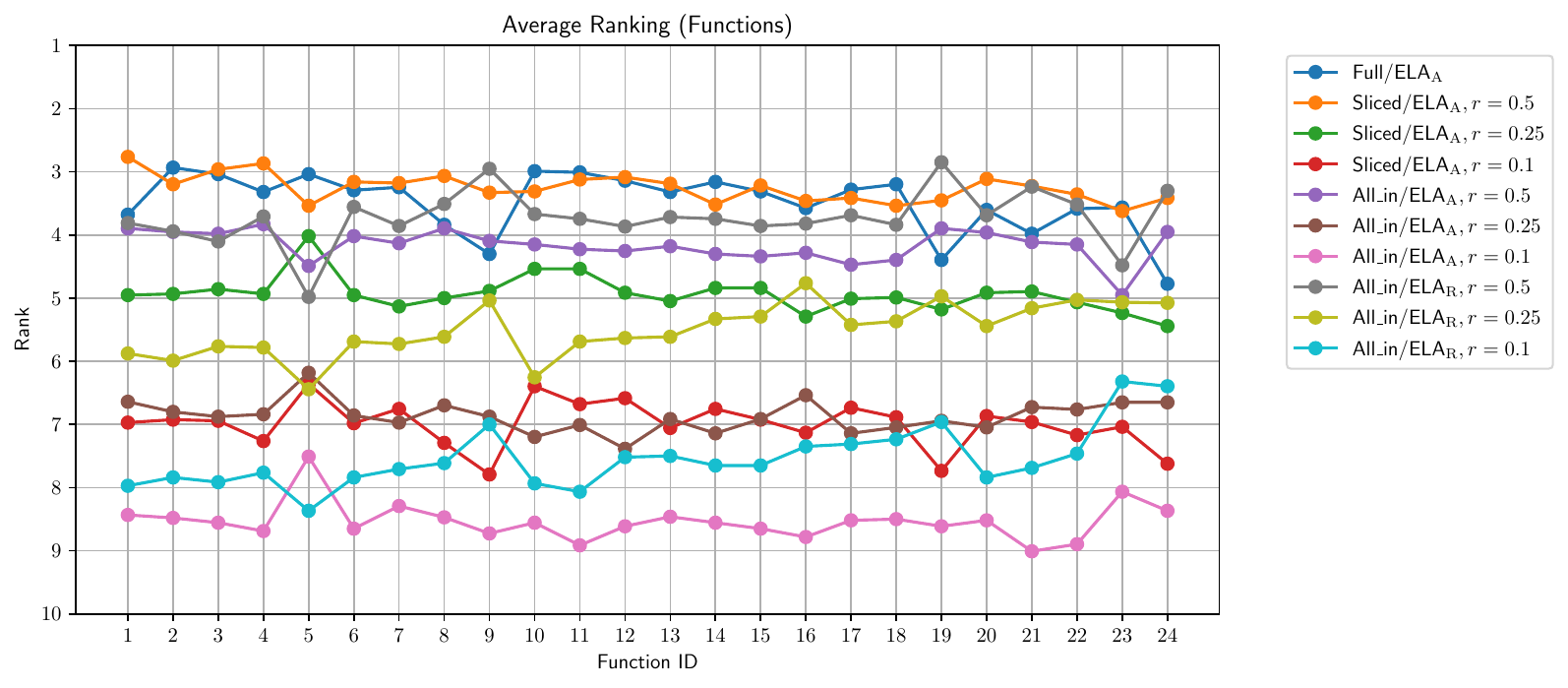}
        \caption{Average method rankings based on the median Wasserstein-1 distance across features per function. 
        }
        \label{fig:wasserstein_rankings_per_function}
    \end{figure}
    \begin{figure}[t]
        \centering
        \includegraphics[width=0.95\linewidth]{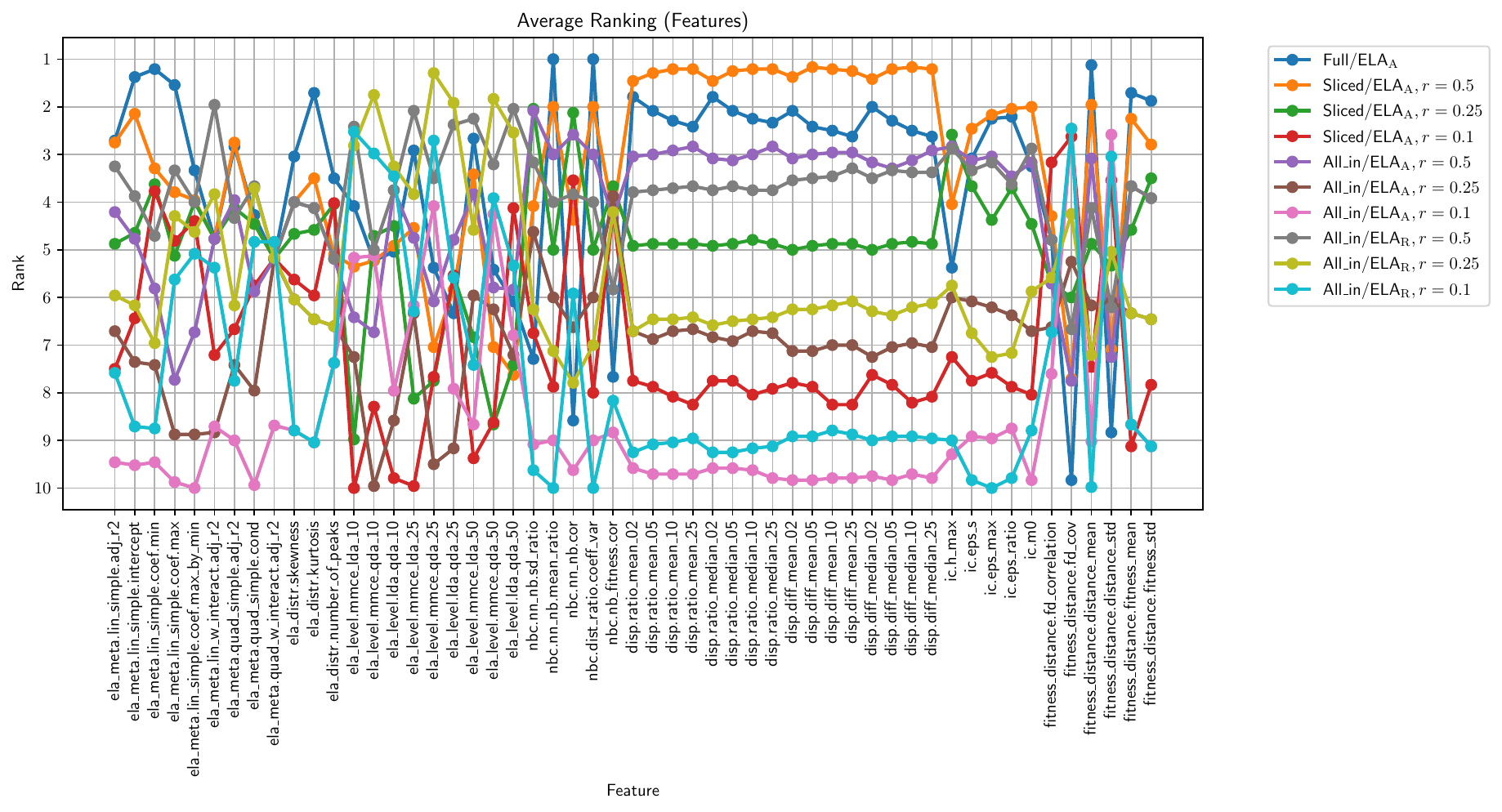}
        \caption{Average method rankings based on the median Wasserstein-1 distance across functions per feature. 
        }
        \label{fig:wasserstein_rankings_per_feature}
    \end{figure}
    Figure~\ref{fig:wasserstein_rankings_per_function} reports the average rankings per function (aggregated over all features), while Figure~\ref{fig:wasserstein_rankings_per_feature} reports the average rankings per feature (aggregated over all functions). 
%
    From Figure~\ref{fig:wasserstein_rankings_per_function}, the methods \texttt{Full/ELA$_{\texttt{A}}$} and \texttt{Sliced/ELA$_{\texttt{A}}$} with \(r = 0.5\) consistently achieve top rankings across functions. In contrast, methods with \(r = 0.1\) systematically rank lowest, indicating that aggressive compression degrades performance. Thus, for the considered feature sets, retaining a moderate level of landscape information is necessary to preserve descriptive power.
    %

    Figure~\ref{fig:wasserstein_rankings_per_feature} shows consistent trends across feature classes. For \texttt{ela\_meta.lin\_simple}, \texttt{Full/ELA$_{\texttt{A}}$} performs best, as these features depend on global coverage for fitting linear meta-models. In contrast, for \texttt{ela\_distr}, \texttt{ic}, and \texttt{disp}, embedding-based sampling with moderate compression ($r = 0.5$) is competitive with or outperforms full-space sampling, indicating that subspace sampling can match or improve upon the default approach.
    %
    For \texttt{ela\_level}, single-embedding strategies combined with feature computation in the reduced space (\emph{e.g.,} \texttt{All\_in/ELA$_{\texttt{R}}$}) achieve the best rankings, suggesting that level-set features benefit from higher sampling density within a single subspace.
    Overall, the effectiveness of a sampling strategy is strongly feature-dependent. 
    However, all these results underscore a key practical implication: the common practice of applying default full-space sampling (\texttt{Full/ELA$_{\texttt{A}}$}) with a limited budget to characterize problem landscapes is often suboptimal.
    Targeted use of structured subspace sampling can meaningfully improve the quality of landscape characterization for specific feature sets.
    
    \paragraph{Overall comparison across features and functions.}
    %
    %
    \begin{figure}[t]
        \centering
        \includegraphics[width=0.975\linewidth]{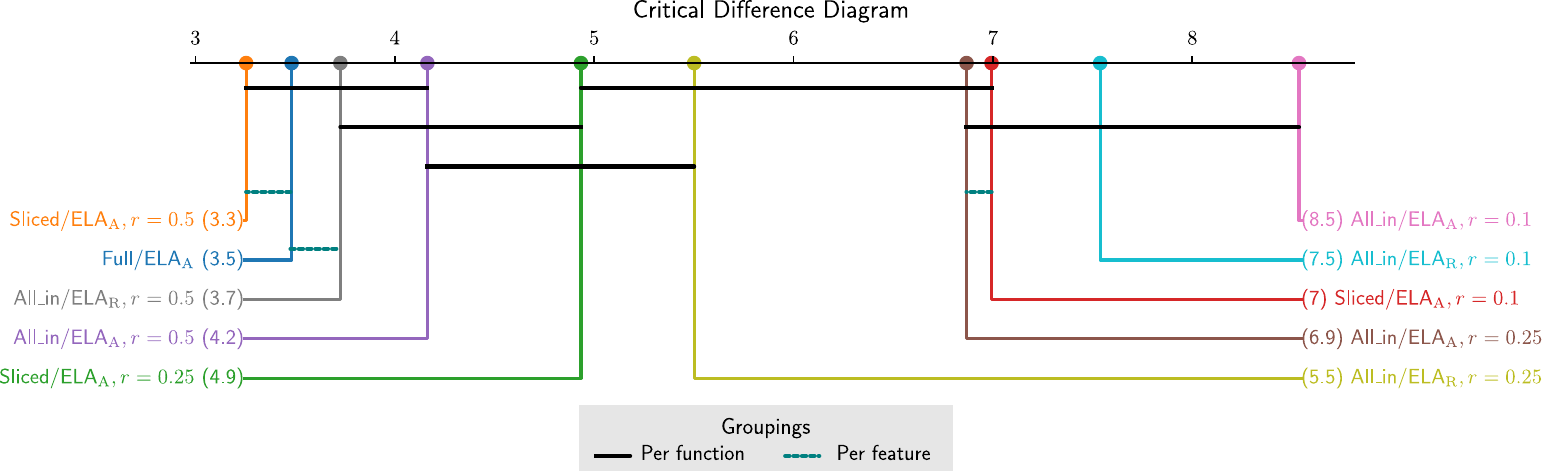}
        \caption{Critical difference diagram with groupings obtained by averaging over all functions and features.}
        \label{fig:cd_diagram_dual}
    \end{figure}
    The CD diagram (Fig.~\ref{fig:cd_diagram_dual}) summarizes the performance of the sampling strategies by aggregating results across both ELA features and problem landscapes. While the overall trends are consistent, the grouping structure reflects the combined variability across descriptors and functions, motivating a cautious interpretation of the aggregated results.
    
    To obtain an overall comparison, we define a consensus ranking by averaging the mean ranks obtained in the feature-wise and function-wise analyses. This criterion assigns equal importance to robustness across features and across functions, and can be interpreted as a compromise between the two views. In addition, we consider the two-dimensional rank space induced by both criteria, where non-dominated methods represent jointly competitive approaches.
    
    According to this consensus criterion, \texttt{Sliced/ELA$_\texttt{A}$} with $r=0.5$ achieves the best overall performance, as it consistently ranks among the top methods in both views. \texttt{Full/ELA$_{\texttt{A}}$} and \texttt{All\_in/ELA$_{\texttt{R}}$} with $r=0.5$ follow closely, forming a group of competitive strategies with only minor differences in average rank. In contrast, methods using lower sampling ratios ($r=0.25$ and $r=0.1$) are consistently dominated, confirming that aggressive reduction of the compression ratio significantly degrades the approximation of feature distributions.
    
    Overall, the results indicate that, despite no single method attains statistically superior results across all considered settings, moderate reduction ratios, when combined with structured sampling strategies, most notably slicing, achieves a favorable trade-off between computational cost and approximation fidelity, leading to consistently strong performance across both ELA features and BBOB functions. 

    \section{Outlook and Conclusion}
    \label{sec:outlook}

In this paper, we investigated sampling in Exploratory Landscape Analysis (ELA) under limited evaluation budgets. Using \textit{random linear embeddings}, we introduced sampling strategies that allocate samples to randomly generated low-dimensional subspaces, increasing sampling density and improving the resolution of local geometric relationships.

Across BBOB functions and feature sets, these strategies are consistently competitive with, and often outperform, full-space sampling under the same budget. The advantage is most pronounced for the feature sets capturing \emph{dispersion} and \emph{information content}. 
Moderate compression ($r = 0.5$) yields the best trade-off, suggesting that concentrating samples within informative subspaces is more effective than distributing them uniformly across the full search space.
These results show that, under limited budgets, sampling is a key design choice in ELA rather than a secondary detail. The standard practice of sampling in the full decision space can therefore be suboptimal and should be reconsidered.
%
%
%
\paragraph{Limitations.}
Our study is restricted to a $20$-dimensional benchmark setting and a specific family of random linear embeddings. The effectiveness of the approach depends on both the problem structure and the choice of subspace dimension: overly low-dimensional slices may fail to capture complex or global landscape properties, and randomly sampled subspaces are not guaranteed to intersect informative regions. 

In addition, not all ELA feature classes benefit equally from subspace-based sampling. In particular, features relying on global structure, such as those from the \textit{meta-model}, \textit{y-distribution}, or parts of the \textit{fitness-distance correlation} sets,
remain difficult to estimate reliably under localized sampling.

These limitations, however, do not alter the main conclusion: under limited budgets, sampling uniformly in the full space should not be taken as the default, and structured subspace sampling provides a viable alternative.

\paragraph{Future Work.}
Several directions follow naturally from this work. 
Our study focuses on BBOB functions, which are not designed to exhibit low intrinsic dimensionality, meaning that relevant variation of the objective function is concentrated in a lower-dimensional subspace. Given that sampling on random embeddings naturally aligns with this setting, future work will consider benchmarks where this property holds, including real-world problems. Another important direction is to assess the impact of these sampling strategies on downstream tasks such as algorithm selection and configuration.

    \begin{credits}

        \subsubsection{\discintname}
        \small{\textbf{The authors have no competing interests to declare that are relevant to the content of this article}}
    \end{credits}
    %
    %
    %
    \bibliographystyle{splncs04}
    \bibliography{references}
    %
    %



\end{document}